\documentclass{article} 
\usepackage{iclr2027_conference,times}

\usepackage{amsmath,amsfonts,bm}

\def\eqref#1{equation~\ref{#1}}

\def\1{\bm{1}}

\DeclareMathAlphabet{\mathsfit}{\encodingdefault}{\sfdefault}{m}{sl}
\SetMathAlphabet{\mathsfit}{bold}{\encodingdefault}{\sfdefault}{bx}{n}

\usepackage{hyperref}
\usepackage{url}

\usepackage{graphicx}
\usepackage{booktabs}
\usepackage{wrapfig}
\usepackage{tcolorbox}
\tcbuselibrary{breakable}
\usepackage{amsmath,amsfonts}
\usepackage{algorithm}
\usepackage{algpseudocode}
\usepackage{array}
\usepackage{textcomp}
\usepackage{pifont}
\usepackage{verbatim}
\usepackage{graphicx}
\usepackage{fancybox}
\usepackage[table]{xcolor}
\usepackage{xcolor}
\usepackage{arydshln}  
\usepackage{booktabs}       
\usepackage{amsfonts}       
\usepackage{nicefrac}       
\usepackage{microtype}      
\usepackage[dvipsnames]{xcolor}         
\usepackage{enumitem}
\usepackage{multirow}
\usepackage{amsmath}
\usepackage{mdframed}
\usepackage{subcaption}
\usepackage{soul}
\usepackage{amssymb}
\usepackage{pifont}

\title{VEvo-Harness: Causally-Guarded Self-Evolvi\\ng Harness for Agentic Image Generation}

\author{%
  Yuan Wang$^{1,2}$, Yuhang Yu$^{1}$, Mengting Chen$^{1,\clubsuit}$, Tianrun Xu$^{2}$, Zihao Pan$^{1}$ \\
  \textbf{Xuetao Feng}$^{2}$, \textbf{Wei Li}$^{1}$ \\
$^{1}$Taobao \& Tmall group of Alibaba \quad
$^{2}$Tsinghua University
\quad $^{\clubsuit}$ Corresponding Author
}
\iclrfinalcopy 
\begin{document}

\maketitle

\begin{abstract}
The central bottleneck in visual generation and editing has migrated from achieving pixel-level fidelity to satisfying densely constrained user specifications through increasingly agentic workflows. Prevailing agentic generation pipelines remain shallow generate-verify-rewrite loops treating VLM judgments as coarse visual feedback, either discarding repair experience or admitting ungated updates that poison skill library. With memorize and govern periods left open, they degrade into ungoverned iterators of repeated trial-and-error rather than continual self-evolving harness. Towards this issue, we present \textbf{VEvo-Harness}, the first causally-guarded continual self-evolving harness system for agentic visual generation. VEvo-Harness introduces four coupled innovations: (1) \textit{a visual traceback} that densifies sparse judgments into root-cause attributions with a hierarchical verification tree; (2) \textit{a visual continual governance} that admits only causally valid skill patches through counterfactual dual-arm image generation replay; (3) \textit{a lifecycle-managed harness state} that hierarchically distills root-cause-anchored episodes across working, experience, and pattern tiers into a persistent skill library of routable and causally verified capabilities; (4) \textit{a heterogeneous sandbox policy} that orchestrates a diverse action space of repair primitives, dynamically routing cost-aware interventions based on root causes to select the most effective strategy. VEvo-Harness surpasses state-of-the-art agentic methods and rivals proprietary generators (\textit{e.g.}, GPT-Image, Nano-Pro) while maintaining the continual harness evolution and generalization.

\end{abstract}

\vspace{-0.7cm}
\section{Introduction}
\vspace{-0.25cm}

Recently, multimodal generation~\citep{song2026joyai,hu2026swift,minimax2026h3,seedance2026seedance,chen2026tstars,liu2026continuous,chen2026orion,chen2024zero} has evolved from research capability into general-purpose utility. Leading proprietary systems such as GPT-Image~\citep{gptimage2-api} and Nano Banana~\citep{gemini2-blog}, alongside prominent open-source frameworks like Qwen-Image~\citep{wu2025qwen} and Z-Image~\citep{cai2025z} achieve unprecedented levels of fidelity. As these models become ubiquitous creative tools, research attention shifts toward increasingly complex prompts. Modern generation requests steadily outgrow a single inference pass, as a growing fraction hinge on exact object counts~\citep{kamath2025geneval}, attribute binding~\citep{zhou2026cpi}, external knowledge~\citep{pan2026wiseedit,wang2026corpora}, or multi-step reasoning~\citep{wang2026expertverse} beyond aesthetic description. The central bottleneck has thus shifted from pixel-level fidelity to satisfying multi-faceted specifications under densely coupled constraints. 

To counter this escalating complexity, recent work shifts scaling from model weights to test-time computation, \textit{i.e.}, prompt densification~\citep{wang2024divide,yang2026dda} and semantic rewriting~\citep{zhan2024prompt,yang2024idea2img}. However, prompt-only interventions lack fine-grained attribution and pixel-grounded feedback, while newer multi-agent systems~\citep{feng2026gen,chen2026genevolve,chen2026genrouter,he2026gems} surround generator with planning, web search, skill invocation, and refinement loops. Despite this progress, their harnesses function as read-only control-flow orchestrators. GEMS and Gen-Searcher augment generation via iterative verification or external retrieval, yet their stateless or statically curated harnesses discard repair trajectories and freeze capability maps at deployment. GenRouter evolves only bucket-level routing utilities while leaving the routed repair loops and their admission criteria untouched, so none of them possesses a governed write path through which verified experience become persistent capability. Moreover, sparse visual feedback makes such naive correlation-based accumulation self-deceiving, inducing harness-level hallucination. We thus ask \textit{how the scaffolding of a frozen visual generator can consolidate cross-task repair experience into persistent state under a causally governed write path, without collapsing into such hallucination}?

\begin{figure}[]
    \centering
    \includegraphics[width=1\textwidth,height=0.53\textwidth]{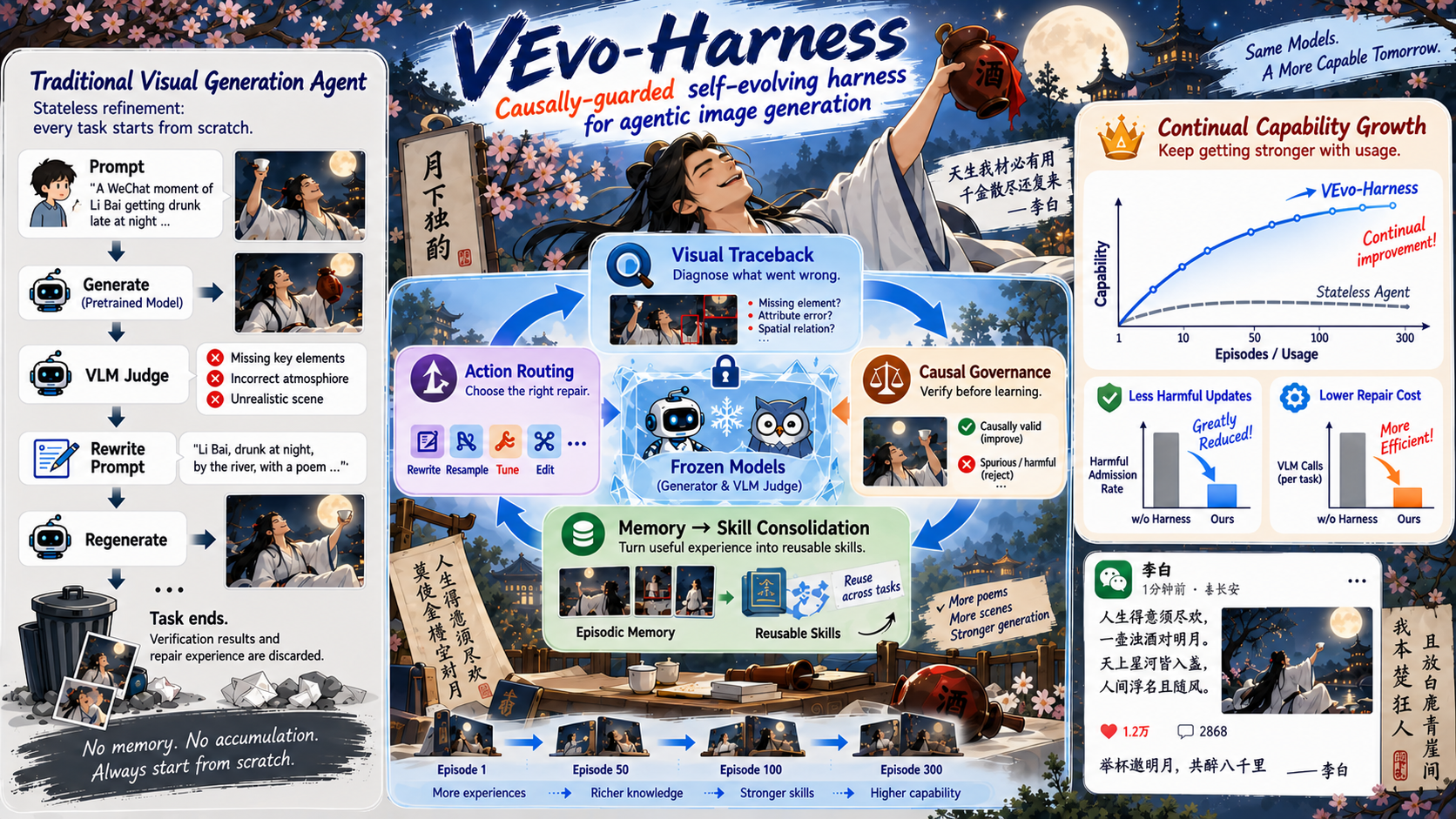}
        \vspace{-0.6cm}
    \caption{
    \textit{From stateless refinement to continual capability growth}. VEvo-Harness causally verifies and reuses visual experience, enabling frozen generators to become progressively stronger with usage.
    }
    \label{fig:teaser}
    \vspace{-0.65cm}
\end{figure}

The emerging agent harness~\citep{kang2026harness,karten2026continual} treats the scaffold around models as a site of continual adaptation to accumulate capability. Extending this to visual synthesis, however, demands an explicit causal guarantee. In Fig.~\ref{fig:teaser}, we present \textbf{VEvo-Harness}, to our knowledge the first continual causally-guarded self-evolving harness for agentic image generation with four principles:

(1) \textbf{Visual traceback.} Unlike coding harnesses~\citep{deepseek-harness2026,openclaw2026} with line-level tracebacks, visual critics provide only coarse binary feedback (\textit{e.g.}, pass/fail), because visual failures are continuous and multi-scaled. We construct a hierarchical verification tree that dynamically expands failed judgments into fine-grained failure modes (\textit{e.g.}, attribute binding), yielding structured root-cause signals that anchor experience memory and causally verified skill updates. (2) \textbf{Visual continual governance}. To prevent harness hallucination, we introduce a visual continual evaluator that estimates each candidate patch's causal effect via counterfactual dual-arm replay under pinned skill routing, and admits only patches with positive estimated effects, thereby keeping the skill library causally validated. (3) \textbf{A lifecycle-managed harness state.} We organize persistent memory into working, instance, and pattern tiers, distilling root-cause-anchored repair episodes into reusable failure patterns that are causally verified before promotion into routable skills. The lifecycle enables the harness to accumulate causally verified experience and self-evolve monotonically across tasks. (4) \textbf{A heterogeneous sandbox policy.} Rather than hard-wiring the harness to a single prompt-rewriting tool, we expand its action registry into a heterogeneous sandbox, and route failures with a cost-aware policy based on visual traceback and persistent harness state. Our approach contributes threefold:

\vspace{-0.2cm}
\begin{itemize}
    \item We propose \textbf{VEvo-Harness}, the first causally-guarded self-evolving harness for agentic image generation. It recasts the scaffold around a frozen generator into a persistent, causally verified state space, establishing a new paradigm of harness-level capability accumulation.
    \item VEvo-Harness integrates four innovations, \textit{i.e.}, counterfactual dual-arm replay for visual continual governance, hierarchical verification for precise failure attribution, lifecycle-managed state for experience accumulation, and sandbox routing policy for adaptive intervention.
    \item Extensive experiments across holistic generation and editing benchmarks show that VEvo-Harness continuously self-evolves via accumulated experience, yielding reset-free monotonic capability growth and zero-shot generalization to highly complex visual instructions.
\end{itemize}


\vspace{-0.5cm}
\section{Related Works}
\label{Related Works}
\vspace{-0.3cm}

\textbf{Agentic Image Generation.} Recent work reframes image generation from text-to-pixel decoding as an agentic runtime, replacing prompt expansion and heuristic rewriting with planning, retrieval, verification, and iterative repair. Qwen-Image-Agent~\citep{zhang2026qwen} casts generation as context construction, while Mind-Brush~\citep{he2026mind}, Gen-Searcher~\citep{feng2026gen}, and RS-Gen~\citep{bian2026rs} use search and reasoning to ground knowledge-intensive prompts. CRAFT~\citep{kovalev2025craft} and SCOPE~\citep{ren2026scope} move toward structured control by maintaining explicit visual constraints and invoking targeted repair only when violations arise. GenClaw~\citep{ye2026genclaw} adds executable visual code as an intermediate canvas for layout control, whereas AR-RAG~\citep{qi2026ar} performs autoregressive patch-level retrieval during decoding. GEMS~\citep{he2026gems} and GenRouter~\citep{chen2026genrouter} extend this toward agent-native runtimes with persistent memory and adaptive workflow routing. They remain tied to fixed workflow families, leaving fine-grained visual attribution and harness-level governance of heterogeneous visual failures insufficiently addressed. Our VEvo-Harness, a unified visual generation scaffold, fill this gap via causally guarded self-evolution.

\textbf{Agentic Harnesses and Scaffolding.} Recent work treats harness as runtime infrastructure rather than prompt wrapper. \cite{ning2026code} identify tool orchestration, state management, and execution verification as core harness functions, and \cite{karten2026prime} emphasizes long-horizon performance depends on persistent substrates, \textit{i.e.}, continual memory, durable state, reusable skills, and runbook control. \cite{lou2026autoharness,zhang2026self,lee2026recursive,pan2026evolving} show  substantial gains come from synthesizing, revising, or optimizing harness itself through environment feedback, regression-tested edits, or interface-side adaptation. \cite{thangarajah2026dcas} further clarifies that scaffold-specific planning conventions can hinder cross-scaffold transfer. Further extensions cover behavior localization, lifecycle governance, runtime mutation, and structured policy-environment interfaces \cite{wang2026harness,garralda2026governed,hou2026harness,gan2026mag}. Prior work focuses on coding, search, and web agents, leaving visual generation underexplored, where sparse feedback and stochastic execution demand causal tracing and persistent adaptation.

\begin{figure}[]
    \centering
    \includegraphics[width=1\textwidth,height=0.42\textwidth]{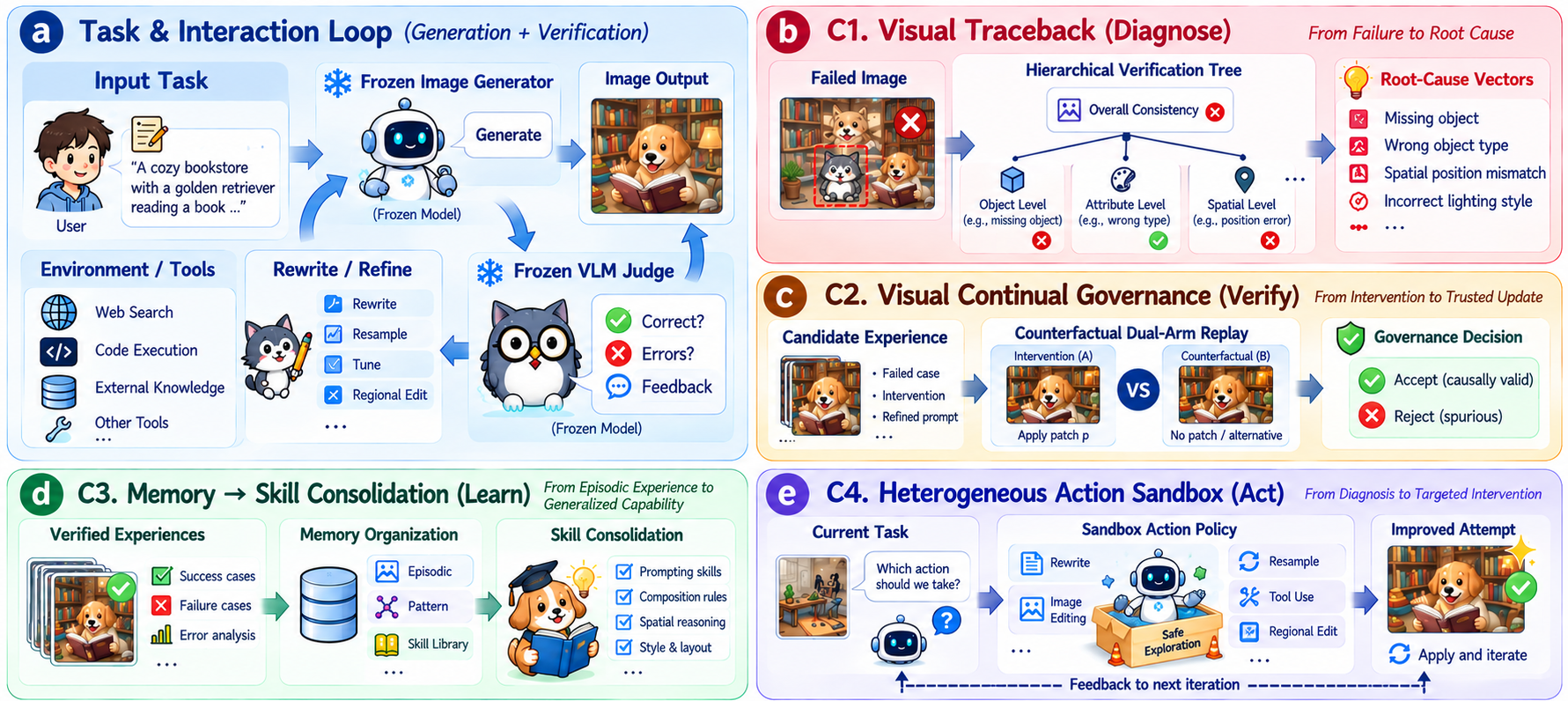}
        \vspace{-0.4cm}
    \caption{Overview of our proposed VEvo-Harness pipeline. A self-evolving scaffold is built around (a) a Interaction Loop for generation and verification, comprising (b) Visual Traceback, (c) Visual Continual Governance, (d) Lifecycle-Managed Harness State, and (e) Heterogeneous Action Sandbox.
    }
    \label{fig:framework}
    \vspace{-0.4cm}
\end{figure}

\vspace{-0.5cm}
\section{Method}
\label{Method}
\vspace{-0.35cm}

\subsection{Preliminary and Framework Overview}
\label{Preliminary and Framework Overview}
\vspace{-0.2cm}

\textbf{Formulation of Agentic Visual Generation.} Let $\mathcal{X}$ and $\mathcal{Y}$ denote the multimodal input conditions and generated image spaces. Agent-based visual generation expands standard single-step inference by introducing a foundational primitive space and orchestrating structured workflows. The generation process is formulated as $y\!=\!M(x, v_{\text{ref}}; \mathcal{H})$, where $v_{\text{ref}}\!\in\!\mathcal{Y}$ represents optional reference images, $M$ is a frozen generation model, and $\mathcal{H}$ denotes the agent harness, formalized as a composite system $\mathcal{H}\!=\!(\mathcal{A}, \mathcal{M}, \mathcal{G}, \pi)$. Here, $\mathcal{A}$ is the execution sandbox, $\mathcal{G}$ is the governance mechanism, $\pi$ is the routing policy, and $\mathcal{M}\!=\!(\mathcal{E}, \mathcal{S})$ denotes the lifecycle-managed state, comprising the episodic memory $\mathcal{E}$ and the persistent skill library $\mathcal{S}$. To handle complex visual synthesis, the harness orchestrates a directed workflow $w\!\in\!\mathcal{W}$ over a primitive set $\mathcal{O}$. In VEvo-Harness, accumulated experience continually updates $\mathcal{H}_t$, forming a self-evolving agent $\mathcal{A}_t\!=\!(M,\mathcal{H}_t,w)$ that generates image $y\!=\!\mathcal{A}_t(x,v_{\text{ref}})$.

\textbf{Framework Overview.} As illustrated in Fig.~\ref{fig:framework}(a) and Algo.1, VEvo-Harness forms a bidirectional closed loop between a frozen generation model $M$ and an evolving harness $\mathcal{H}_t$, which integrates external tools (\textit{e.g.}, web search, code execution) to drive iterative generation via targeted interventions. In the forward flow, a user prompt is first parsed into verifiable visual constraints. The routing policy $\pi$ then queries the \textbf{lifecycle-managed harness state} (Sec.~\ref{Lifecycle-Managed Harness State}) $\mathcal{M}$ to retrieve failure-aware episodic traces from $\mathcal{E}$ and applicable heuristics from $\mathcal{S}$, which guide workflow construction and prompt enhancement. In the backward flow, failures of generated images are localized by \textbf{visual traceback} (Sec.~\ref{Visual Traceback}) and addressed via the \textbf{heterogeneous sandbox policy} (Sec.~\ref{Heterogeneous Sandbox Policy}). The repaired trajectories are assessed by \textbf{visual continual governance} (Sec.~\ref{Visual Continual Governance}), which promotes only causally validated gains into the persistent skill library and retains the others as episodic experience. With generate--verify--intervene--memorize loop, the harness continually evolves through recursive self-improvement.

\vspace{-0.4cm}
\subsection{Visual Traceback}
\label{Visual Traceback}
\vspace{-0.3cm}

Agentic visual generation relies on a judge VLM to evaluate a set of compositional visual constraints $c$—spanning object counting, attribute binding, and spatial relations—against a synthesized image $y_{\text{cand}}$, yielding a sparse Boolean verdict vector $\mathbf{v}\!\in\!\{0,1\}^m$, where $v_j\!=\!\text{VLM}(y_{\text{cand}}, c_j)$. Crucially, these constraints are cached per prompt to ensure strict criterion alignment across all arms of the downstream counterfactual protocol. While sufficient for accept/reject decisions, binary signals obscure visual failure attributions (\textit{e.g.}, spatial distortion or attribute misbinding), offering no actionable, pixel-aligned guidance for targeted intervention when $v_j\!=\!0$. As shown in Fig.~\ref{fig:framework}(b), we introduce the \textit{Visual Traceback} mechanism, which densifies these non-differentiable global verdicts into structured, semantically grounded root causes, enabling precise diagnostic feedback for downstream repair.

\vspace{-0.15cm}
Visual traceback densifies each failed boolean verdict into localized root causes via a per-constraint two-level verification tree $\mathcal{T}_j$, whose root encodes the coarse requirement $c_j$ and expands lazily only when $v_j\!=\!0$. Upon failure, the VLM is prompted to generate a bounded set of finer-grained sub-questions that disentangle the failure mechanism. This generation is explicitly guided to span orthogonal diagnostic dimensions—encompassing object existence, fine-grained attributes, and stylistic fidelity—while semantically redundant sub-questions are dynamically filtered. Unlike a predefined dimension checklist, this \textit{failure-conditioned adaptive decomposition} tailors the diagnostic structure to the specific semantic nature of $c_j$ (\textit{e.g.}, relational constraints expand into spatial and existence checks). All sub-questions of all failed constraints are evaluated concurrently, bounding the diagnostic overhead to a single parallelized VLM round. The visual traceback process is defined as:
\begin{equation}
\setlength{\abovedisplayskip}{1pt}
\setlength{\belowdisplayskip}{1pt}
    \mathbf{e} = \text{Traceback}(\mathbf{v}, y_{\text{cand}}, c) = \bigcup\nolimits_{j:v_j=0} \mathrm{Leaves}^{-}(\mathcal{T}_j),
\end{equation}
where $\mathrm{Leaves}^{-}(\mathcal{T}_j)$ collects false leaf sub-questions, or $c_j$ if none exist. Thus, $\mathbf{e}$ converts a scalar verdict into re-verifiable, pixel-grounded visual predicates actionable by subsequent sandbox policy.

\vspace{-0.45cm}
\subsection{Lifecycle-Managed Harness State}
\label{Lifecycle-Managed Harness State}
\vspace{-0.3cm}

Conventional harnesses suffer from limited transfer in flat episodic buffers or degradation in append-only skill registries. As illustrated in Fig.~\ref{fig:framework}(d), we thus architect a lifecycle-managed harness state $\mathcal{M}\!=\!\{\mathcal{M}_{\text{work}}, \mathcal{E}, \mathcal{P}, \mathcal{S}\}$ with a three-tier distillation pipeline. Working memory $\mathcal{M}_{\text{work}}$ buffers per-iteration prompts, pixel-level verifications, and structured root-cause vectors of the current task, flushing at task completion to prevent context leakage. Experience memory $\mathcal{E}$ archives trajectories that generate images passing compositional verification, enabling retrieval of refined prompts that previously resolved similar failures. Pattern memory $\mathcal{P}$ distills recurring visual failure modes into preventive strategies injected before rendering, shifting the cost from post-hoc repair to pre-emptive prevention. $\mathcal{S}$ consumes causally verified knowledge from this pipeline as routable, executable skills.

\vspace{-0.15cm}
Each entry $\tau_j\!=\!(\mathbf{e}_j, \xi_j, p_j^{\ast}, \Delta_{\text{cf}}^{(j)})$ in $\mathcal{E}$ explicitly anchors a verified repair to its structured root-cause vector $\mathbf{e}_j$ rather than to raw prompt-image pairs, with $\xi_j$, $p_j^{\ast}$, and $\Delta_{\text{cf}}$ denoting the refinement sequence, verified prompt, and counterfactual causal utility (Sec.~\ref{Visual Continual Governance}). Two-stage retrieval first recalls relevant candidates by root-cause similarity $r(\mathbf{e}_{\text{curr}}, \tau_j)\!=\!\mathrm{sim}(\mathbf{e}_{\text{curr}}, \mathbf{e}_j)$ to match current failure mode, then reranks them by descending $\Delta_{\text{cf}}$ to return intervention with the highest validated causal gain.

\vspace{-0.1cm}
As repair trajectories accumulate in $\mathcal{E}$, an induction operator distills recurring visual faults into pattern memory $\mathcal{P}$. Each pattern $\rho_k\!=\!(\mu_k, \phi_k, \sigma_k, \kappa_k)$ abstracts a cluster of trajectories sharing similar root-cause vectors. $\mu_k$ denotes the generalized failure mode, $\phi_k$ the preventive strategy, $\sigma_k$ the empirical support count, and $\kappa_k$ the dynamic confidence score. The confidence $\kappa_k$ is modulated by usage feedback, which tracks the empirical reliability of the preventive strategy $\phi_k$ (\textit{details in Appendix}).
Ineffective patterns are progressively pruned, leaving only high-signal regularities in $\mathcal{P}$. Promotion to the skill library $\mathcal{S}$ is strictly gated by counterfactual verification. Post-admission, verified patches undergo conflict-aware consolidation via Exponential Moving Average (EMA), which permits state updates only when a new patch $\mathbf{s}_{\text{new}}$ satisfies semantic and causal criteria:
\begin{equation}
    \setlength{\abovedisplayskip}{2pt}
    \setlength{\belowdisplayskip}{2pt}
    \text{Merge}(\mathbf{s}_{\text{new}}) \iff \text{Sim}(\phi_{\mathbf{s}_{\text{new}}}, \phi_{\mathbf{s}_{\text{exist}}}) > \theta_{\text{sem}} 
    \land \ \Delta_{\text{cf}}(\mathbf{s}_{\text{new}}) > \Delta_{\text{cf}}(\mathbf{s}_{\text{exist}}),
    \label{eq:skill_merge}
\end{equation}
where $\phi_{\mathbf{s}}$ is the preventive strategy's semantic embedding, which prevents redundancy while ensuring causal superiority over rendered images. When merged, the existing strategy is updated via $\phi_{\text{exist}}\!\leftarrow\!\alpha \phi_{\text{exist}}\!+\!(1-\alpha)\phi_{\text{new}}$ to balance historical stability with new evidence. The usage feedback reinforces or penalizes retrieved patterns based on pixel-level verifications. To prevent memory bloat, a capacity-aware eviction policy prunes elements via a composite score of confidence decay and idleness:
\begin{equation}
    \setlength{\abovedisplayskip}{2pt}
    \setlength{\belowdisplayskip}{2pt}
    \text{Evict}(x_k) = \arg\min\nolimits_{x \in \mathcal{P} \cup \mathcal{S}} [ \lambda_1\!\cdot\!\kappa_x + \lambda_2\!\cdot\!\text{idle}(x) + \lambda_3\!\cdot\!\text{fail}(x) ],
    \label{eq:memory_eviction}
\end{equation}
where $x_k$ represents either a pattern $\rho_k\!\in\!\mathcal{P}$ or a skill $\mathbf{s}_k\!\in\!\mathcal{S}$, $\text{idle}(x)$ is the time since last reinforcement, and $\text{fail}(x)$ is the recent failure count. Elements falling below a threshold $\theta_{\text{prune}}$ are purged. Success preservation protects validated rules from later overwrites, while $\pi$ jointly retrieves skills and root-cause evidence from $\mathcal{E}, \mathcal{P}$ to condition generation with verified instructions and episodic context. 

\begin{figure}[]
    \centering
    \vspace{-0.25cm}
    \includegraphics[width=1\textwidth,height=0.53\textwidth]{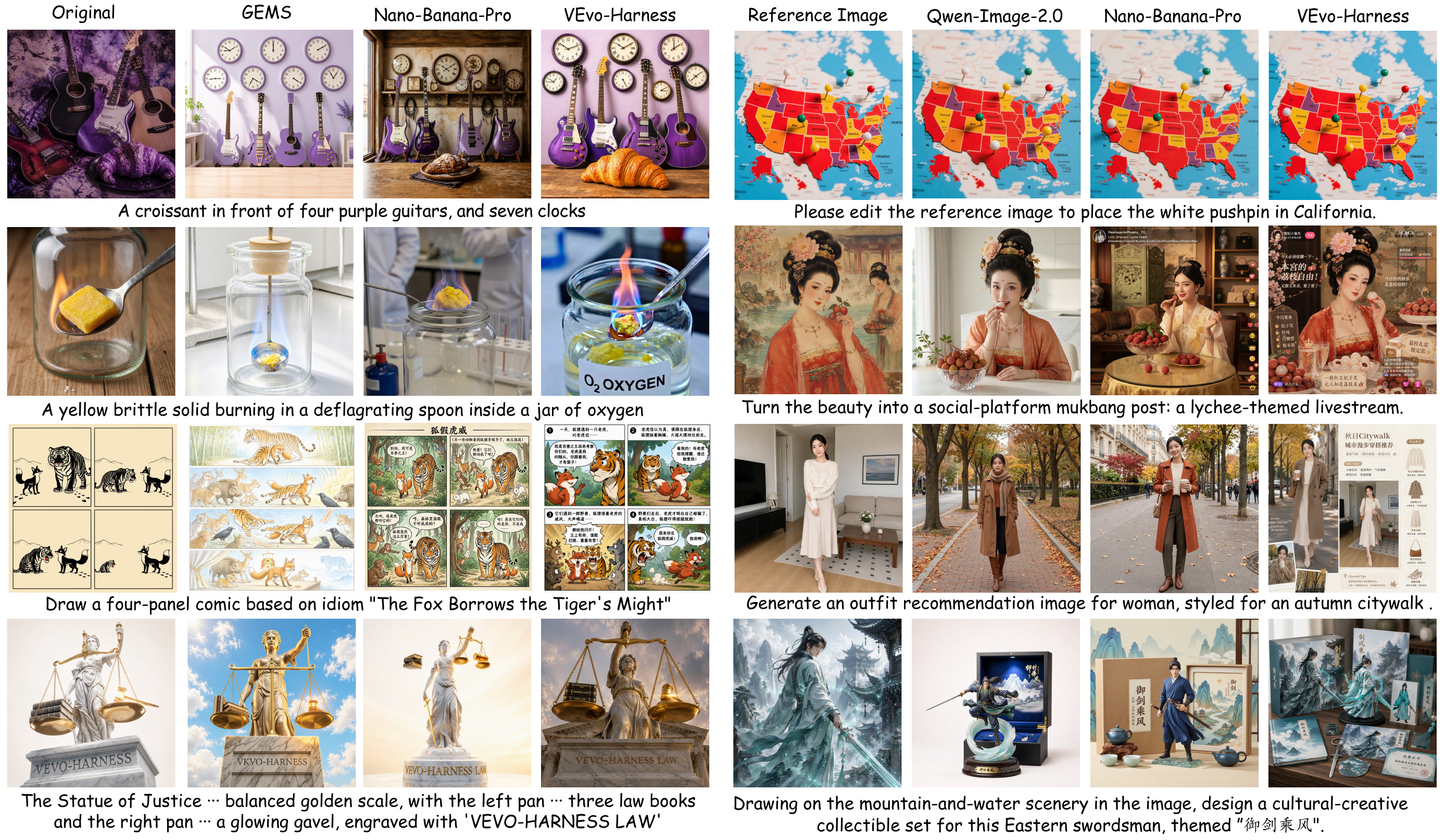}
        \vspace{-0.45cm}
    \caption{Qualitative results of VEvo-Harness with baselines on generation (\textit{left}) and editing (\textit{right}).}
    \label{fig:Qualitative_results}
    \vspace{-0.5cm}
\end{figure}

\vspace{-0.47cm}
\subsection{Visual Continual Governance}
\label{Visual Continual Governance}
\vspace{-0.3cm}

Traditional harness updates based on text-level success rates, risking harness-level hallucination where textually plausible edits degrade visual fidelity. To address this, we propose Visual Continual Governance in Fig.~\ref{fig:framework}(c), a causal admission protocol that certifies every candidate skill patch by counterfactually replaying it against the unpatched skill on rendered pixels so that only interventions with positive visual causal utility enter the skill library. Let $\mathcal{G}(\cdot)$ be the frozen generator and $\mathcal{R}_{\mathrm{vis}}(\cdot)$ the reward verifying the image $y$ against visual contract $\mathcal{C}_x$. The causal effect of a candidate patch $\phi$ on skill $s_k$ is intervention $s_k \oplus \phi$, whose counterfactual utility is expected visual reward shift over $z$:
\begin{equation}
\setlength{\abovedisplayskip}{1pt}
\setlength{\belowdisplayskip}{1pt}
    \Delta_{\mathrm{cf}}(\phi \mid s_k, x) = \mathbb{E}_{z \sim p(\mathcal{Z})} \Big[ \mathcal{R}_{\mathrm{vis}}\big(\mathcal{G}(x, s_k \oplus \phi; z), \mathcal{C}_x\big) - \mathcal{R}_{\mathrm{vis}}\big(\mathcal{G}(x, s_k; z), \mathcal{C}_x\big) \Big].
    \label{eq:causal_utility_theory}
\end{equation}
After computing the causal utility $\Delta_{\mathrm{cf}}$, we apply a three-valued admission gate: verified ($\Delta_{\mathrm{cf}}\!>\!\epsilon$), detrimental ($\Delta_{\mathrm{cf}}\!<\!0$), or uncertain ($0\!<\!\Delta_{\mathrm{cf}}\!<\!\epsilon$). Only verified patches are incorporated into the skill library. To avoid repeatedly attempting rejected edit directions, we introduce a denied-change buffer $\mathcal{D}$ that archives detrimental and uncertain patches. 
Proposals exhibiting semantic equivalence to any denied edits, quantified by $\max_{\phi'\!\in\!\mathcal{D}}\!\mathcal{J}(\phi, \phi')\!>\!\theta_{\mathcal{J}}$, are pre-emptively suppressed prior to the verification. It precludes redundant verification of rejected failure patterns. Post-admission, patches incorporated into skill library $\mathcal{S}$ undergo continuous monitoring to prevent the catastrophic forgetting. For each skill $s_k\!\in\!\mathcal{S}$, we track per-sample records indexed by $(s_k, x)$ and compute pass ratio shifts $\delta_r(x)\!=\! r_t(x)\!-\!r_{t-1}(x)$ at epoch boundaries. When a run satisfies visual contract, correctly rendered instruction fragments are stored as protected rules $\psi^+$ in $s_k$, preventing later failure-driven updates from degrading verified content while repairing other defects. As patches accumulate, the skill library risks unbounded growth and conflicting visual priors. We regulate these updates via semantic conflict resolution and edit-budget scheduling to prevent visual degradation (\textbf{see Appendix~\ref{Data, Metrics, and Implementation Details} for details}).


\vspace{-0.5cm}
\subsection{Heterogeneous Sandbox Policy}
\label{Heterogeneous Sandbox Policy}
\vspace{-0.3cm}

Existing pipelines~\citep{he2026gems,feng2026gen} adopt a monolithic intervention paradigm that attributes all visual failures to prompt misalignment, disregarding heterogeneous causal origins. To this issue, as shown in Fig.~\ref{fig:framework}(e), we construct a heterogeneous action sandbox $\mathcal{A}\!=\!\{a_{\text{rewrite}}, a_{\text{resample}}, a_{\text{tune}}\}$ comprising distinct intervention types, each action intervening on a specific component of the image generation pipeline. Given the structured root-cause vector $\mathbf{e}$ from visual traceback (Sec.~\ref{Visual Traceback}), the routing policy $\pi$ selects actions via a softmax distribution parameterized by risk-sensitive Q-values:
\begin{equation}
\setlength{\abovedisplayskip}{1pt}
\setlength{\belowdisplayskip}{1pt}
    \pi(a \mid \mathbf{e}) = \exp\left(\hat{Q}(a, \mathbf{e}) / \tau\right)/{\sum\nolimits_{a' \in \mathcal{A}} \exp\left(\hat{Q}(a', \mathbf{e}) / \tau\right)}, \quad \hat{Q}(a, \mathbf{e}) = \hat{\mu}(a, \mathbf{e}) - \lambda \cdot \hat{\sigma}^2(a, \mathbf{e}),
    \label{eq:policy_distribution}
\end{equation}
where $\hat{\mu}(a, \mathbf{e})$ and $\hat{\sigma}^2(a, \mathbf{e})$ are the empirical mean and variance of $\Delta_{\mathrm{cf}}(a)$ accumulated over historical rollouts, $\tau$ is a temperature parameter controlling the exploration-exploitation tradeoff, and $\lambda \geq 0$ penalizes high-variance actions. The Q-value estimates are updated via EMA after each rollout: 
\begin{equation}
\setlength{\abovedisplayskip}{2pt}
\setlength{\belowdisplayskip}{2pt}
    \hat{\mu}(a, \mathbf{e}) \leftarrow (1-\alpha)\hat{\mu}(a, \mathbf{e}) + \alpha \cdot \Delta_{\mathrm{cf}}(a), \quad \hat{\sigma}^2(a, \mathbf{e}) \leftarrow (1-\alpha)\hat{\sigma}^2(a, \mathbf{e}) + \alpha \left(\Delta_{\mathrm{cf}}(a) - \hat{\mu}(a, \mathbf{e})\right)^2,
    \label{eq:bayesian_update}
\end{equation}
By matching each failure mode to its structurally appropriate intervention, the heterogeneous sandbox policy achieves targeted repair at the true causal origin, broader failure coverage across generation pipeline levels, and adaptive convergence toward consistently effective intervention strategies.

\newcommand{\up}[1]{\,{\color{green!60!black}$\uparrow${\normalsize #1}}}
\newcommand{\dn}[1]{\,{\color{red}$\downarrow${\normalsize #1}}}

\begin{table}[]
\vspace{-0.2cm}
\caption{Main results on general image generation benchmarks. All inference-time scaling methods are evaluated on the same frozen backends, where \textbf{bold} and \underline{underlined} mark the best and second-best.}
\vspace{-0.21cm}
\label{tab:image_generation_benchmarks}
\centering
\resizebox{\columnwidth}{!}{
\begin{tabular}{ccccccccc}
\toprule
\multicolumn{2}{c}{\textbf{Model}} & \textbf{GenEval} & \textbf{GenEval2} & \textbf{DPG-Bench} & \textbf{OneIG-EN} & \textbf{OneIG-ZH} & \textbf{WISE} & \textbf{Avg.} \\ \hline
\multicolumn{2}{c}{Nano-Banana~\cite{comanici2025gemini}}   & 0.75 & 44.6 & 85.23 & 0.550 & 0.405 & -- & -- \\
\multicolumn{2}{c}{GPT-Image-1~\cite{hurst2024gpt}}   & 0.84 & --   & 85.15 & 0.533 & 0.474 & 0.80 & -- \\
\multicolumn{2}{c}{Seedream-4~\cite{seedream2025seedream}}    & 0.84 & --   & 88.25 & 0.576 & 0.553 & 0.78 & -- \\ \midrule
\multicolumn{2}{c}{Bagel~\cite{deng2025emerging}}         & 0.82 & 23.1 & 85.10 & 0.361 & 0.370 & 0.70 & 59.35 \\
\multicolumn{2}{c}{Ovis-U1~\cite{wang2025ovis}}         & 0.89 & 21.7 & 83.72 & 0.382 & 0.395 & 0.64 & 56.02 \\
\multicolumn{2}{c}{Echo-4o~\cite{ye2025echo}}         & 0.89 & 67.9 & 86.07 & 0.467 & 0.488 & 0.74 & 68.75 \\
\multicolumn{2}{c}{Z-Image-Turbo~\cite{cai2025z}} & 0.77 & 31.0 & 85.08 & 0.526 & 0.501 & 0.57 & 60.29 \\
\multicolumn{2}{c}{Qwen-Image~\cite{wu2025qwen}}    & 0.87 & 33.8 & 88.32 & 0.539 & 0.548 & 0.62 & 65.09 \\ \midrule
\multicolumn{9}{c}{\textit{Inference-Time Scaling}} \\ 
 & Original~\citep{cai2025z}  & 0.77\up{0.00} & 31.0\up{0.0} & 85.08\up{0.00} & 0.526\up{0.000} & 0.501\up{0.000} & 0.57\up{0.00} & 60.29 \\
 & Rewrite~\citep{he2026gems}   & 0.77\up{0.00} & 39.7\up{8.7} & 84.48\dn{0.60} & 0.520\dn{0.006} & 0.469\dn{0.032} & 0.84\up{0.27} & 66.93 \\
 & Promptist~\citep{hao2023optimizing} & 0.76\dn{0.01} & 32.3\up{1.3} & 65.69\dn{19.39} & 0.333\dn{0.193} & 0.327\dn{0.174} & 0.54\dn{0.03} & 52.20 \\
 & Search~\citep{ma2025inference}    & 0.81\up{0.04} & 43.5\up{12.5} & 85.47\up{0.39} & 0.530\up{0.004} & 0.506\up{0.005} & 0.61\up{0.04} & 64.55 \\
 & Maestro~\citep{wan2025maestro}   & 0.82\up{0.05} & 44.6\up{13.6} & 85.29\up{0.21} & 0.548\up{0.022} & 0.519\up{0.018} & 0.85\up{0.28} & 70.05 \\
 & CRAFT~\citep{kovalev2025craft}     & 0.80\up{0.03} & 62.4\up{31.4} & 85.29\up{0.21} & 0.582\up{0.056} & 0.542\up{0.041} & 0.78\up{0.21} & 72.38 \\
 & GEMS~\citep{he2026gems}      & 0.86\up{0.09} & 63.5\up{32.5} & 86.01\up{0.93} & 0.569\up{0.043} & 0.552\up{0.051} & 0.81\up{0.24} & 74.51 \\
 & GenRouter~\citep{chen2026genrouter}      & —— & 57.6\up{26.6} & 85.49\up{0.41} & 0.570\up{0.044} & 0.545\up{0.044} & 0.86\up{0.29} & 68.12 \\ 
\multirow{-9}{*}{\rotatebox{90}{Z-Image-Turbo}} & VEvo-Harness (Ours) & \textbf{0.89}\up{0.12} & \textbf{67.8}\up{36.8} & \textbf{86.72}\up{1.64} & \textbf{0.594}\up{0.068} & \textbf{0.563}\up{0.062} & \textbf{0.88}\up{0.31} & \textbf{76.28} \\ \midrule
 & Original~\citep{wu2025qwen}  & 0.66\up{0.00} & 29.0\up{0.0} & 84.69\up{0.00} & 0.487\up{0.000} & 0.489\up{0.000} & 0.59\up{0.00} & 57.50 \\
 & Rewrite~\citep{he2026gems}   & 0.68\up{0.02} & 39.8\up{10.8} & 83.49\dn{1.20} & 0.465\dn{0.022} & 0.435\dn{0.054} & \underline{0.81}\up{0.22} & 63.46 \\
 & Promptist~\citep{hao2023optimizing} & 0.66\up{0.00} & 27.4\dn{1.6} & 64.02\dn{20.67} & 0.288\dn{0.199} & 0.307\dn{0.182} & 0.57\dn{0.02} & 48.83 \\
 & Search~\citep{ma2025inference}    & 0.74\up{0.08} & 42.1\up{13.1} & 86.17\up{1.48} & 0.502\up{0.015} & 0.497\up{0.008} & 0.67\up{0.08} & 63.84 \\
 & Maestro~\citep{wan2025maestro}   & 0.76\up{0.10} & 48.0\up{19.0} & 84.07\dn{0.62} & 0.494\up{0.007} & 0.489\up{0.000} & 0.84\up{0.25} & 68.24 \\
 & CRAFT~\citep{kovalev2025craft}     & 0.79\up{0.13} & 66.3\up{37.3} & 85.87\up{1.18} & 0.533\up{0.046} & 0.518\up{0.029} & 0.79\up{0.20} & 72.54 \\
 & GEMS~\citep{he2026gems}      & 0.79\up{0.13} & 70.4\up{41.4} & 85.59\up{0.90} & 0.542\up{0.055} & 0.532\up{0.043} & 0.80\up{0.21} & 73.74 \\
 & GenRouter~\citep{chen2026genrouter}      & —— & 71.6\up{42.6} & 87.39\up{2.70} & 0.553\up{0.066} & 0.541\up{0.052} & \textbf{0.88}\up{0.29}  & 71.28 \\ 
\multirow{-9}{*}{\rotatebox{90}{Qwen-Image-2512}} & VEvo-Harness (Ours) & \textbf{0.90}\up{0.24} & \textbf{86.1}\up{57.1} & \textbf{88.14}\up{3.45} & \textbf{0.572}\up{0.085} & \textbf{0.581}\up{0.092} & 0.86\up{0.27} & \textbf{77.59}  \\ \bottomrule
\end{tabular}
}
\vspace{-0.65cm}
\end{table}

\vspace{-0.44cm}
\section[Experiments]{Experiments\footnote{Cost-efficiency trade-offs, framework details, holistic ablations, and visualizations are in Appendix.}}
\vspace{-0.3cm}

\subsection{Experimental Setup}
\vspace{-0.15cm}

\textbf{Implementation Details.} We evaluate VEvo-Harness on Z-Image-Turbo and Qwen-Image-2512, using Qwen3.5-27B for the core reasoning backend, with at most 5 harness iterations. For visual traceback, each failed constraint expands up to 4 sub-questions judged by Gemini-3.7-Flash. For visual continual governance, $\Delta_{\mathrm{cf}}$ uses $K\!=\!4$ seed-fixed rollouts per arm; the denied buffer holds 50 edits, suppresses proposals with $\mathcal{J}\!>\!0.6$, and refines verified patches within $\operatorname{sim}\!=\!0.45$. The edit budget cosine-decays from $\eta_0\!=\!4$ to $\eta_{\min}\!=\!1$ over 20 steps, with slow updates every 25 runs. For sandbox policy, $(\tau,\lambda,\alpha,W)\!=\!(1.0,0.5,0.3,5)$. Lifecycle consolidation starts after 10 episodes and 5 runs, distilling up to 5 patterns from the latest 30 episodes; patterns merge at Jaccard $>0.6$, update confidence by $\pm0.05$, decay by 0.02, and are pruned below 0.15. Retrieval uses the top-3 patterns and top-2 instances; skill merging triggers at $|\Phi_k|\!>\!8$ with $\theta_{\mathrm{sem}}\!=\!0.6$, and EMA coefficient $\alpha\!=\!0.3$.

\vspace{-0.15cm}
\textbf{Benchmarks and Baselines.} Following the protocol of GEMS~\citep{he2026gems}, we assess the VEvo-Harness on five general image generation benchmarks—GenEval~\citep{ghosh2023geneval}, GenEval2~\citep{kamath2025geneval}, DPG-Bench~\citep{hu2024ella}, OneIG~\citep{chang2026oneig}, and WISE~\citep{niu2025wise}—and extend the analysis to downstream tasks (\textit{e.g.}, creativity, spatial reasoning) on LongText-Bench~\citep{geng2025x}, SpatialGenEval~\citep{wang2026everything}, CREA~\citep{venkatesh2026crea}, and ArtiMuse~\citep{cao2026artimuse}. We also incorporate four editing benchmarks: GEdit-Bench~\citep{liu2025step1x}, RISE-Bench~\citep{zhao2026envisioning}, KRIS-Bench~\citep{wu2026kris}, WiseEdit~\citep{pan2026wiseedit}. 

\begin{table}[]
\vspace{-0.2cm}
\caption{Generalization results to downstream tasks comparison with inference-time scaling methods.}
\vspace{-0.21cm}
\label{tab:downstream}
\centering
\resizebox{\columnwidth}{!}{
\begin{tabular}{cccccccc}
\toprule
\multicolumn{2}{c}{\textbf{Model}} & \textbf{LongText-EN} & \textbf{LongText-ZH} & \textbf{SpatialGenEval} & \textbf{CREA} & \textbf{ArtiMuse} & \textbf{Avg.} \\ \midrule
 & Original~\citep{cai2025z}  & 0.912\up{0.000} & 0.932\up{0.000} & 58.7\up{0.0} & 11.84\up{0.00} & 43.27\up{0.00} & 58.41 \\
 & Rewrite~\citep{he2026gems}   & 0.571\dn{0.341} & 0.495\dn{0.437} & 59.6\up{0.9} & 16.75\up{4.91} & 57.86\up{14.59} & 56.65 \\
 & Promptist~\citep{hao2023optimizing} & 0.034\dn{0.878} & 0.070\dn{0.862} & 48.3\dn{10.4} & 14.00\up{2.16} & 43.74\up{0.47} & 35.98 \\
 & Search~\citep{ma2025inference}    & 0.918\up{0.006} & 0.937\up{0.005} & 59.3\up{0.6} & 12.98\up{1.14} & 44.21\up{0.94} & 59.88 \\
 & Maestro~\citep{wan2025maestro}  & 0.877\dn{0.035} & 0.807\dn{0.125} & 60.3\up{1.6} & 15.81\up{3.97} & 56.86\up{13.59} & 63.52 \\
 & CRAFT~\citep{kovalev2025craft}     & 0.951\up{0.039} & 0.760\dn{0.172} & 60.6\up{1.9} & 13.63\up{1.79} & 54.95\up{11.68} & 61.63 \\
 & GEMS~\citep{he2026gems} & \underline{0.952}\up{0.040} & \underline{0.940}\up{0.008} & \underline{61.4}\up{2.7} & \underline{22.55}\up{10.71} & \underline{58.58}\up{15.31} & \underline{72.44} \\
\multirow{-8}{*}{\rotatebox{90}{Z-Image-Turbo}} & VEvo-Harness (Ours) & \textbf{0.961}\up{0.049} & \textbf{0.948}\up{0.016} & \textbf{62.8}\up{4.1} & \textbf{25.73}\up{13.89} & \textbf{60.42}\up{17.15} & \textbf{75.62} \\ \midrule
 & Original~\citep{cai2025z}  & 0.854\up{0.000} & 0.892\up{0.000} & 60.0\up{0.0} & 17.45\up{0.00} & 59.94\up{0.00} & 66.35 \\
 & Rewrite~\citep{he2026gems}   & 0.563\dn{0.291} & 0.468\dn{0.424} & 60.3\up{0.3} & 18.12\up{0.67} & 62.23\up{2.29} & 58.62 \\
 & Promptist~\citep{hao2023optimizing} & 0.030\dn{0.824} & 0.058\dn{0.834} & 48.7\dn{11.3} & 18.35\up{0.90} & 60.65\up{0.71} & 43.73 \\
 & Search~\citep{ma2025inference}    & 0.883\up{0.029} & 0.907\up{0.015} & 60.6\up{0.6} & 18.71\up{1.26} & 59.96\up{0.02} & 68.11 \\
 & Maestro~\citep{wan2025maestro}  & 0.833\dn{0.021} & 0.792\dn{0.100} & 61.0\up{1.0} & 17.78\up{0.33} & 61.62\up{1.68} & 65.78 \\
 & CRAFT~\citep{kovalev2025craft}     & 0.897\up{0.043} & 0.787\dn{0.105} & 61.6\up{1.6} & 15.61\dn{1.84} & 59.78\dn{0.16} & 64.40 \\
 & GEMS~\cite{he2026gems} & \underline{0.913}\up{0.059} & \underline{0.931}\up{0.039} & \underline{62.1}\up{2.1} & \underline{24.01}\up{6.56} & \underline{62.89}\up{2.95} & \underline{74.31} \\
\multirow{-8}{*}{\rotatebox{90}{Qwen-Image-2512}} & VEvo-Harness (Ours) & \textbf{0.926}\up{0.072} & \textbf{0.944}\up{0.052} & \textbf{63.5}\up{3.5} & \textbf{27.68}\up{10.23} & \textbf{64.75}\up{4.81} & \textbf{77.93} \\ \bottomrule
\end{tabular}
}
\vspace{-0.61cm}
\end{table}

\subsection{Main Results on Image Generation and Editing Benchmarks}
\vspace{-0.15cm}

\textbf{Performance on Image Generation Benchmarks.} Tab.~\ref{tab:image_generation_benchmarks} shows that VEvo-Harness achieves the best average performance on both backends, reaching 76.28\% on Z-Image-Turbo~\citep{cai2025z} and 77.59\% on Qwen-Image-2512~\citep{wu2025qwen}. Gains are strongest on compositionally demanding benchmarks: on GenEval2, VEvo-Harness reaches 86.1\% with Qwen-Image-2512, outperforming GenRouter~\citep{chen2026genrouter} by 14.5\%, while Promptist~\citep{hao2023optimizing} degrades by 19.3\% on DPG-Bench~\citep{hu2024ella} under Z-Image-Turbo. VEvo-Harness also leads on OneIG~\citep{chang2026oneig} (0.594/0.572) and WISE~\citep{niu2025wise} (0.88/0.86), indicating root-cause decomposition, causal skill verification, and sandbox routing improve both compositional fidelity, retrieval-augmented synthesis, and knowledge-intensive reasoning. By distilling failure/success experiences into verified patterns, VEvo-Harness enables reliable knowledge reuse that improves cross-task generations. With the scaffolding of VEvo-Harness, Qwen-Image surpasses commercial Seedream-4~\citep{seedream2025seedream} on GenEval (0.90), releasing more scaling potential than single-loop and routing-only optimization.

\vspace{-0.15cm}
\textbf{Performance on Downstream Tasks.} The downstream tasks in Tab.~\ref{tab:downstream} verify that capability growth resides purely in agent harness, with VEvo-Harness improving average by 17.21\% and 11.58\% over the original backends. Inference-time scaling methods, \textit{e.g.}, Maestro improve certain metrics but degrade others, revealing that unguided search or experience accumulation triggers performance trade-offs across diverse constraints. VEvo-Harness avoids this by closing generate-verify-intervene-memorize loop around a persistent state space. It distills domain-specific preventive patterns for text rendering, spatial reasoning, and aesthetic fidelity from root-cause-anchored episodes, while counterfactual dual-arm replay further gates every patch before deployment keeping skill library causally validated. VEvo-Harness achieves state-of-the-art results, elevating CREA by 13.89\% and 10.23\%, and ArtiMuse by 17.15\% and 4.81\% on two backends. These gains show that accumulated harness state transfers zero-shot to specialized domains, confirming reset-free growth at harness layer. 

\vspace{-0.15cm}
\textbf{Performance on Image Editing Benchmarks.} As shown in Tab.~\ref{tab:image_editing_benchmarks}, VEvo-Harness establishes state-of-the-art results among community models, surpassing proprietary systems on Temporal (RISE), Conceptual (KRIS), and GEdit. On RISE-Bench, it reaches 45.8 on Temporal and 20.0 on Logical, exceeding Know-Thinker by 1.1\% and 4.7\%, benefiting from the visual traceback and intervention routing. On KRIS-Bench, which spans factual, conceptual, and procedural knowledge, it attains 84.93, 84.29, and 85.26, outperforming Nano-Banana-Pro on Conceptual, since the lifecycle-managed state reuses distilled knowledge patterns across tasks. On WiseEdit, covering awareness, interpretation, and imagination, VEvo-Harness reaches 80.6, and scores 8.63 on GEdit, surpassing all community baselines. Success preservation protects validated source-image predicates, preventing identity drift and background degradation during localized edits. VEvo-Harness leads all knowledge-intensive metrics by up to 2.0\%, showing causally verified harness evolution transfers effectively to editing.

\vspace{-0.15cm}
\textbf{Visual Comparison.} Fig.~\ref{fig:Qualitative_results} reports visualizations of VEvo-Harness and leading proprietary/agentic systems on generation and editing tasks. VEvo-Harness consistently advances complex infographic generation, precise text rendering, visual aesthetics, and knowledge-intensive edit. With scaffolded loop, VEvo-Harness generalizes to structurally dense visual instructions, confirming that reusable lifelong memory extend internal generation priors beyond the native capacity of frozen backbones.

\begin{table*}[t]
\centering
\vspace{-0.25cm}
\caption{Ablation studies of VEvo-Harness. (a) Component-level ablation analysis of Evo-Harness. (b) Micro-ablation of intra-module details, with VLM calls and wall-clock latency reported as cost.}
\vspace{-0.3cm}
\begin{subtable}[t]{0.52\textwidth}
\centering
\label{tab:ablation_all}
\caption{Component-level ablation of the VEvo-Harness system.}
\label{tab:ablation_component}
\small
\setlength{\tabcolsep}{0.6pt}
\begin{tabular}{l c c c c}
\toprule
\textbf{Variant} & \textbf{G-Eval2} & \textbf{Wise} & \textbf{W-Edit} & \textbf{Kris} \\
\midrule
\multicolumn{5}{l}{\textit{Component Ablation}} \\
\textit{Lifecycle-Managed State} \\
\quad Stateless Harness     & 79.4 & 0.81 & 76.2 & 78.6 \\
\quad w/o skill library $\mathcal{S}$    & 83.6 & 0.84 & 79.1 & 81.5 \\
\textit{Visual Continual Governance} \\
\quad Admit-All (no causal gate)         & 83.2 & 0.84 & 78.4 & 81.6 \\
\quad Text-Only Gate       & 84.5 & 0.85 & 79.3 & 82.8 \\
\textit{Visual Traceback} \\
\quad Flat Binary Verdict                & 81.8 & 0.83 & 77.5 & 80.2 \\
\quad Fixed-Dimension Checklist          & 83.9 & 0.85 & 79.0 & 82.1 \\
\textit{Heterogeneous Sandbox Policy} \\
\quad Rewrite-Only (monolithic)          & 82.4 & 0.82 & 78.0 & 80.8 \\
\quad Uniform Random Routing             & 84.1 & 0.84 & 79.2 & 82.4 \\
\textbf{Full VEvo-Harness (Ours)}        & \textbf{86.1} & \textbf{0.86} & \textbf{80.6} & \textbf{84.8} \\
\midrule
\multicolumn{5}{l}{\textit{Progressive Assembly}} \\
\quad Base harness                       & 77.6 & 0.78 & 74.2 & 76.9 \\
\quad + Visual Traceback                 & 79.4 & 0.80 & 75.6 & 78.4 \\
\quad + Heterogeneous Sandbox            & 81.2 & 0.82 & 77.1 & 79.8 \\
\quad + Lifecycle-Managed State          & 83.4 & 0.84 & 78.9 & 82.0 \\
\quad + Visual Cont. Governance & \textbf{86.1} & \textbf{0.86} & \textbf{80.6} & \textbf{84.8} \\
\bottomrule
\end{tabular}
\end{subtable}
\hfill
\begin{subtable}[t]{0.44\textwidth}
\centering
\caption{The design-detail micro-ablation. VLM calls and wall-clock latency per task reported as cost.}
\label{tab:ablation_detail}
\vspace{-0.15cm}
\small
\setlength{\tabcolsep}{1.3pt}
\begin{tabular}{l c c c}
\toprule
\textbf{Design detail} & \textbf{Avg.} & \textbf{Calls} & \textbf{Lat.(s)} \\
\midrule
\multicolumn{4}{l}{\textit{Visual Traceback}} \\
\quad w/o lazy expansion & 84.10 & 11.8 & 40.1 \\
\quad w/o redundancy filtering           & 83.62 & 6.6 & 39.2 \\
\quad w/o root fallback                  & 82.94 & 6.2 & 39.6 \\
\multicolumn{4}{l}{\textit{Visual Cont. Governance}} \\
\quad w/o denied-change buffer           & 83.12 & 9.4 & 41.3 \\
\quad w/o success preservation           & 81.63 & 6.4 & 43.8 \\
\quad binary admission gate              & 81.90 & 6.5 & 42.5 \\
\multicolumn{4}{l}{\textit{Lifecycle-Managed State}} \\
\quad w/o working memory  & 83.31 & 6.8 & 40.6 \\
\quad w/o instance memory   & 80.42 & 8.1 & 45.2 \\
\quad w/o pattern memory   & 82.27 & 7.3 & 43.1 \\
\quad lexical retrieval & 80.87 & 7.9 & 44.0 \\
\quad w/o $\Delta_{\text{cf}}$ reranking & 82.58 & 6.9 & 41.7 \\
\quad w/o conf. decay   & 83.47 & 6.4 & 40.2 \\
\multicolumn{4}{l}{\textit{Heterogeneous Sandbox}} \\
\quad w/o variance penalty  & 82.41 & 7.4 & 42.8 \\
\quad w/o EMA Q-update                   & 81.76 & 7.8 & 43.4 \\
\quad w/o tune action                    & 82.13 & 6.7 & 41.9 \\
\textbf{Full VEvo-Harness}        & \textbf{84.38} & \textbf{6.2} & \textbf{38.4} \\
\bottomrule
\end{tabular}
\end{subtable}
\end{table*}

\vspace{-0.1cm}
\begin{figure}[]
    \centering
    \includegraphics[width=1\textwidth,height=0.265\textwidth]{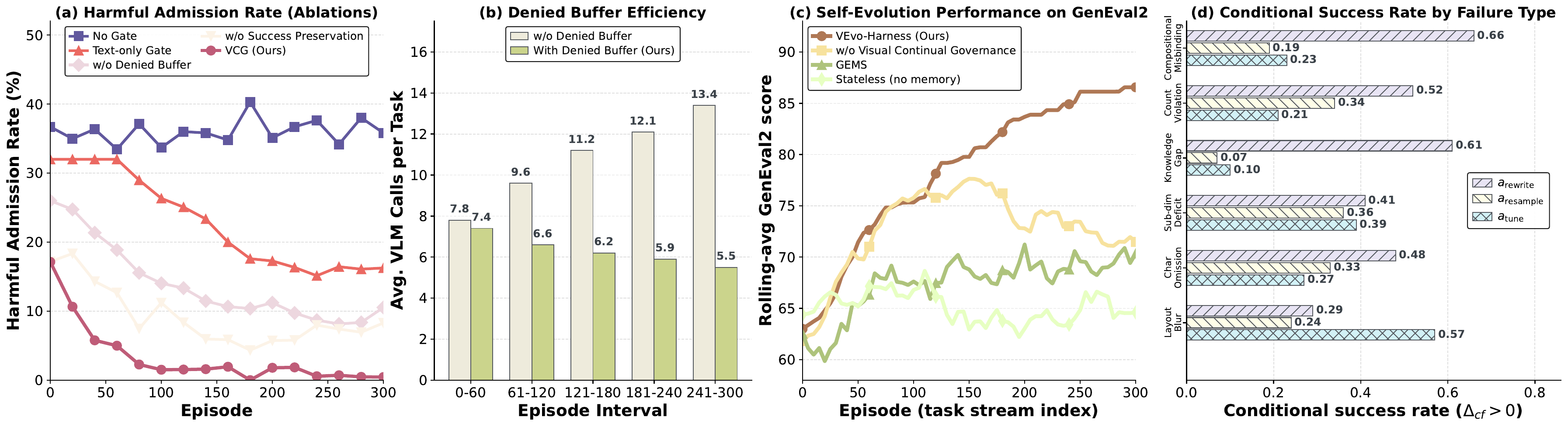}
        \vspace{-0.6cm}
    \caption{\textbf{Governance and Intervention Analysis.}
    (a) Harmful admission rate under governance. (b) Per-task verification cost w/. and w/o. denied-change buffer. (c) Continuous self-evolution results comparisons on GenEval2. (d) Conditional success rate of each sandbox action per failure type.
    }
    \label{fig:compard_with_others}
    \vspace{-0.41cm}
\end{figure}

\begin{table}[h]
\centering
\vspace{-0.35cm}
\caption{Evaluation on instruction-driven editing benchmarks, where \textbf{bold} indicates the best results.}
\vspace{-0.35cm}
\label{tab:image_editing_benchmarks}
\resizebox{\columnwidth}{!}{
\setlength{\tabcolsep}{4.5pt}
\begin{tabular}{cccccccccc}
\toprule
\multirow{2}{*}{\textbf{Model}} & \multicolumn{4}{c}{\textbf{RISE-Bench}} & \multicolumn{3}{c}{\textbf{KRIS-Bench}} & \multirow{2}{*}{\textbf{WiseEdit}} & \multirow{2}{*}{\textbf{GEdit}} \\ 
 & Temp. & Causal & Spatial & Logical & Factual & Concept. & Proced. & & \\ \midrule
Seedream-4.0~\citep{seedream2025seedream}   & 12.9 & 12.2 & 11.0 & 7.1  & 39.26 & — & — & 75.2 & 7.68 \\
Nano-Banana~\citep{gemini2-blog}         &  25.9  & 47.8 & 37.0 & 18.8 & 80.48 & 78.11 & 75.31 & 75.0 & 6.51 \\
Nano-Banana-Pro~\citep{nanobananapro}       & 41.2 & 61.1 & 48.0 & 37.6 & 86.36 & 84.03 & 85.31 & 82.4 & 7.73 \\
GPT-Image-1~\citep{hurst2024gpt}    & 34.1 & 32.2 & 37.0 & 10.6 & 79.80 & 81.37 & 78.32 & 77.6 & 7.53 \\ \midrule
OmniGen~\citep{xiao2025omnigen}        & 1.2 & 1.0  & 0.0  & 1.2  & 33.11 & 28.02 & 23.89 & 36.6 & 5.00 \\
EMU2~\citep{sun2024generative}           & 1.2 & 1.1  & 0.0  & 0.0  & 45.40 & 37.54 & 34.91 & 38.3 & 5.41 \\
Step1X-Edit~\citep{liu2025step1x}    & 0.0 & 2.2  & 2.0  & 3.5  & 45.52 & 48.01 & 31.82 & 49.5 & 6.44 \\
FLUX.1~\citep{batifol2025flux}        & 2.3 & 5.5  & 13.0 & 1.2  & 53.28 & 50.36 & 42.53 & 43.2 & 6.51 \\
UniWorld~\citep{lin2025uniworld}       & — & —   & —   & —   & 47.71 & 44.80 & 47.92 & 49.4 & 7.83 \\
Bagel (w/CoT)~\citep{deng2025emerging}  & 5.9 & 17.8 & 21.0 & 1.2  & 66.18 & 61.92 & 49.02 & 60.0 & 6.52 \\
Uni-CoT~\citep{qin2025uni}        & 8.2 & 18.9 & 20.0 & 1.2  & 71.85 & 71.85 & 63.68 & 60.1 & 6.74 \\
Qwen-Image-Edit~\citep{wu2025qwen}  & 21.2 & 18.9 & 31.0 & 4.7  & 75.89 & 61.15 & 67.55 & 60.2 & 7.88 \\
Know-Thinker~\citep{wang2026expertverse}   & 44.7  & \textbf{52.2} & \textbf{46.0} & 15.3 & 82.53 & 82.65 & 83.32 & 77.3 & — \\ 
VEvo-Harness (Ours)    &  \textbf{45.8} & 47.7 & 37.0 & \textbf{20.0} & \textbf{84.93} & \textbf{84.29} & \textbf{85.26} & \textbf{80.6} & \textbf{8.63} \\ \bottomrule
\end{tabular}
}
\vspace{-0.3cm}
\end{table}

\vspace{-0.3cm}
\subsection{Ablation Study and Discussion}
\vspace{-0.25cm}

\textbf{Component Effectiveness and Execution Efficiency.} Tab.~\ref{tab:ablation_component} identifies lifecycle-managed state as the main driver of self-evolution, as removing persistent memory drops GenEval2/KRIS by 6.7\%/6.2\% and disabling skill consolidation still costs 2.5\%/3.3\%, confirming that retained experience compounds only after distillation into reusable skills. Counterfactual governance beats Admit-All by 2.9\%/3.2\% and exceeds the fixed checklist of Text-Only Gate by 2.2\%/2.7\% on all benchmarks, validating image-based verification before skill updates. HSP improves over Rewrite-Only by 4.0\% on WISE and over random routing by 2.4\% on KRIS, while progressive assembly raises the marginal GenEval2 gain from 1.8\% to 2.7\%, showing the complementary value of action diversity and adaptive routing. Tab.~\ref{tab:ablation_detail} further shows that removing root-cause-anchored instance memory causes the largest degradation (-3.96\%) with latency rising to 45.2s, counterfactual reranking contributes 1.80\%, and EMA updates plus variance regularization add 2.62\% and 1.97\%. On cost, lazy gating cuts VLM calls 47.5\% at a 0.28\% gain, the denied-change buffer blocks redundant replay that would inflate calls 51.6\%, and removing success preservation costs 2.75\% and 43.8s at unchanged calls. The full setting is thus the most accurate (84.38\%), fastest (38.4s), and cheapest (6.2 calls) variant.

\vspace{-0.15cm}
\textbf{Self-evolution via Governed Knowledge Accumulation.} Figs.~\ref{fig:compard_with_others}(a) and (c) link skill-update quality to sustained gains, as VEvo-Harness lifts its rolling-average GenEval2 score from 63.2 to 86.5 by distilling recurring failures into reusable skills. Without visual continual governance (VCG), gains collapse after a mid-stream peak near 78 to about 71.5, ending roughly 15\% below the full model, since unverified updates pollute the skill library. Fig.~\ref{fig:compard_with_others}(a) attributes this collapse to admission quality, where No Gate admits 33--40\% harmful patches and Text-only Gate settles near 16\%, whereas full VCG reaches about 2\% by 100 episodes, rising to 10\% and 8\% without the denied buffer or success preservation. GEMS and the stateless variant finish near 70 and 64.5, confirming that cross-task reuse requires persistent state. Thus lifelong memory and causally-guarded governance are complementary, with persistent state enabling reuse and counterfactual replay filtering updates before propagation.

\vspace{-0.15cm}
\textbf{Governance Efficiency and Redundancy Suppression.} Across 300 streamed episodes, the denied-change buffer progressively reduces verification overhead and harmful admission rates by filtering previously rejected edit directions (Fig.~\ref{fig:compard_with_others}(b)). Without it, repeatedly evaluating structurally similar but previously rejected interventions inflates VLM calls by 71.8\% over the task stream. With the buffer, the call count instead drops by 25.7\%, yielding a 5.1\% reduction in the first interval that widens to 59.0\% by episodes 241--300 and 41.6\% on average (10.82 to 6.32). Fig.\ref{fig:compard_with_others}(a) and the prolonged repair trajectories in Tab.\ref{tab:ablation_detail} show that removing success preservation degrades the admission curve: without shielding correctly rendered elements, the agent repeatedly breaks and re-fixes the same regions. Caching rejected patches further reduces redundant verification and lowers harmful admission rate.

\vspace{-0.15cm}
\textbf{Causal Matching of Heterogeneous Interventions.} Fig.~\ref{fig:compard_with_others}(d) reports $P(\Delta_{cf}\!>\!0)$ for each sandbox action conditioned on failure type, exposing a strongly heterogeneous intervention--response surface. Conditioning-side failures favor \textit{prompt rewriting} policy, with rewrite reaching 0.66 on compositional misbinding, whereas renderer-side defects invert this preference, as the layout blur responds more reliably to hyperparameter tuning (0.57) than to prompt rewriting (0.29). The sub-dim deficit failures do not admit a single dominant action and therefore require stochastic routing over action set. Matched actions outperform mismatched ones by up to \textcolor{green}{\textbf{8.7x}}, indicating that monolithic pipelines structurally bottleneck when a fixed intervention is applied across heterogeneous failure modes, while routing $\pi(a|\mathbf{e})$ over traceback root causes recovers per-type intervention specificity without exhaustive replay.

\begin{wrapfigure}{}{0.5\textwidth}
    \centering
    \vspace{-0.5cm}
    \includegraphics[width=0.5\textwidth]{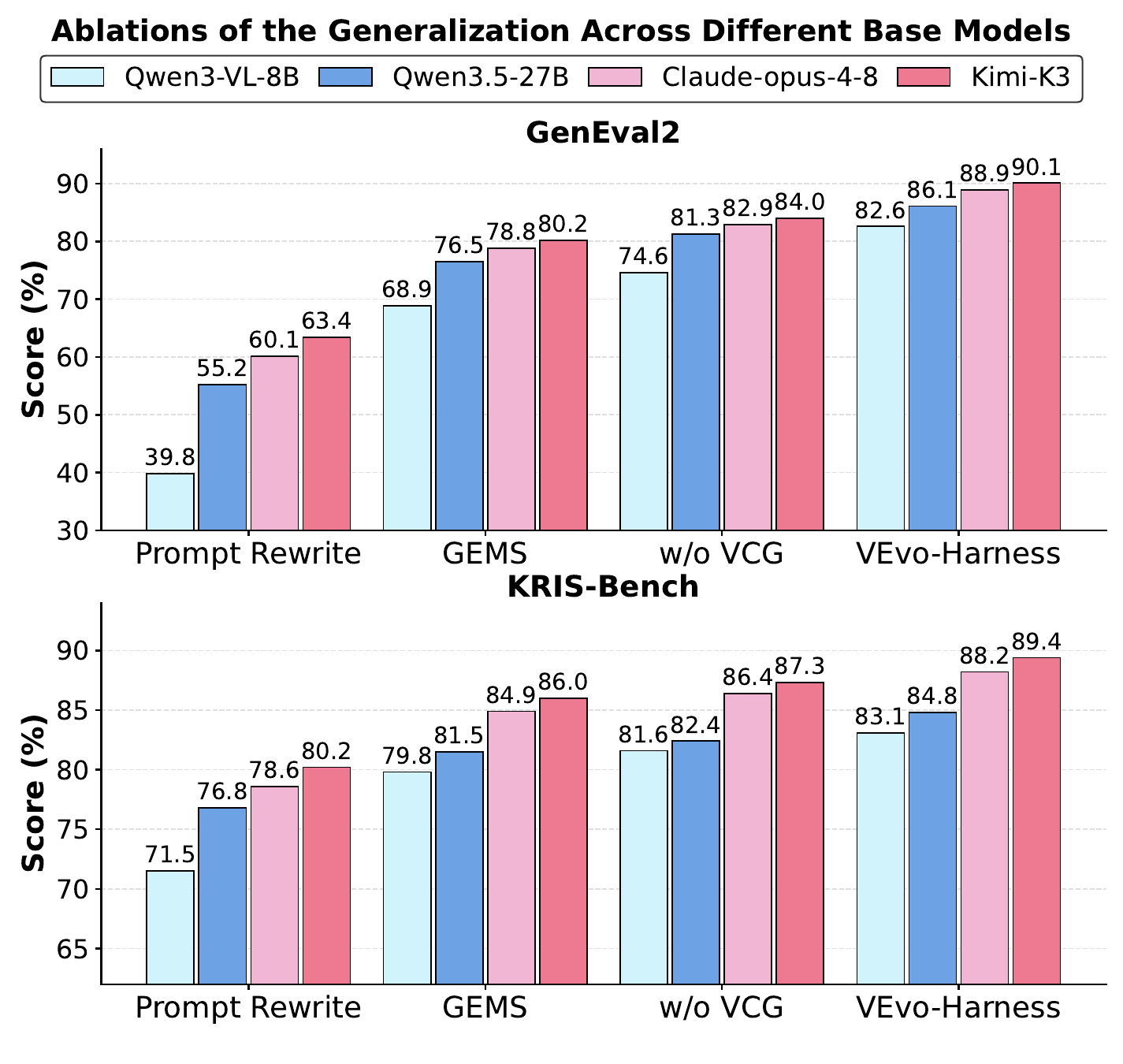}
    \vspace{-0.77cm}
    \caption{Generalization on different base models.}
    \label{fig:generalization}
    \vspace{-0.25cm}
\end{wrapfigure}

\vspace{-0.15cm}
\textbf{Generalization Across Base Models.} Fig.~\ref{fig:generalization} evaluates the transferability of VEvo-Harness across heterogeneous foundation backbones under identical configuration. VEvo-Harness consistently improves both lightweight and frontier models. On GenEval2, Qwen3-VL-8B increases from 74.6\% without the VCG to 82.6\% with full harness, while Kimi-K3~\citep{team2026kimi} improves from 84.0\% to 90.1\%. A similar trend holds on KRIS, where Kimi-K3 rises from 87.3\% to 89.4\%. Significant gains indicate that VEvo-Harness improves a shared visual recovery runtime by self-evolving skills and causal visual governance, rather than model-specific biases. The larger GenEval2 gains further show that root-cause diagnosis and intervention routing transfer effectively across models, supporting VEvo as a robust visual-generation scaffold.

\vspace{-0.25cm}
\section{Conclusion}
\vspace{-0.25cm}

We introduce \textbf{VEvo-Harness}, a reset-free, causally guarded framework elevating the agent scaffold from a transient execution layer to a persistent, self-evolving runtime for agentic image generation. \textit{Visual Traceback} employs a hierarchical verification tree to decompose visual judgments into localized root-cause vectors for fine-grained diagnosis. \textit{Visual Continual Governance} uses counterfactual dual-arm replay to admit only causally validated patches while preserving verified solutions. \textit{Lifecycle-Managed Harness State} structures memory into working, experience, and pattern tiers alongside a patch-evolvable skill library, distilling episodic experiences into reusable capabilities. \textit{Heterogeneous Sandbox Policy} establishes a generalized action routing that dynamically aligns diagnosed root causes with diverse interventions via risk-aware utility estimation. Extensive results across general generation benchmarks confirm that VEvo-Harness achieves substantial gains over agentic pipelines. These results chart a path toward self-evolving agents in e-commerce AIGC workflows (\textit{e.g.}, visual creative generation and virtual try-on), accumulating user-specific abilities through continuous interactions. 

\bibliography{iclr2027_conference}
\bibliographystyle{iclr2027_conference}

\end{document}